\def\year{2021}\relax
\documentclass[letterpaper]{article} % DO NOT CHANGE THIS
\usepackage{aaai21}  % DO NOT CHANGE THIS
\usepackage{times}  % DO NOT CHANGE THIS
\usepackage{helvet} % DO NOT CHANGE THIS
\usepackage{courier}  % DO NOT CHANGE THIS
\usepackage{subfigure}
\usepackage{amsfonts,amssymb}
\usepackage{amsmath}
\usepackage{graphicx}
\usepackage[hyphens]{url}  % DO NOT CHANGE THIS
\usepackage{graphicx} % DO NOT CHANGE THIS
\usepackage{natbib}  % DO NOT CHANGE THIS AND DO NOT ADD ANY OPTIONS TO IT
\usepackage{caption} % DO NOT CHANGE THIS AND DO NOT ADD ANY OPTIONS TO IT
\title{Does Head Label Help for Long-Tailed Multi-Label Text Classification}
\author {
    Lin Xiao\textsuperscript{\rm 1},
    Xiangliang Zhang \textsuperscript{\rm 2},
    Liping Jing \textsuperscript{\rm 1}\thanks{Liping Jing is the corresponding author.},
    Chi Huang \textsuperscript{\rm 1},
    Mingyang Song\textsuperscript{\rm 1}\\
}
\affiliations {
    \textsuperscript{\rm 1}
    Beijing Key Lab of Traffic Data Analysis and Mining,
    Beijing Jiaotong University, Beijing, China \\
    \textsuperscript{\rm 2} King Abdullah University of Science and Technology(KAUST), Saudi Arabia\\
    xiaolin@bjtu.edu.cn,  xiangliang.zhang@kaust.edu.sa,
    lpjing@bjtu.edu.cn, \\
    17271034@bjtu.edu.cn,
    19112026@bjtu.edu.cn

}
\begin{document}
\maketitle

\begin{abstract}
Multi-label text classification (MLTC) aims to annotate documents with the most relevant labels from a number of candidate labels. In real applications, the distribution of label frequency often exhibits a long tail, i.e., a few labels are associated with a large number of  documents (a.k.a. \emph{head} labels), while a large fraction of labels are associated with a  small number of documents (a.k.a. \emph{tail} labels).
To address the challenge of insufficient training data on tail label classification, we propose a \emph{Head-to-Tail Network} (HTTN) to transfer the \emph{meta}-knowledge from the data-rich head labels to data-poor tail labels.
The \emph{meta}-knowledge is the mapping from few-shot network parameters to many-shot network parameters, which aims to promote the generalizability of \emph{tail} classifiers.
Extensive experimental results on three benchmark datasets demonstrate that HTTN consistently outperforms the state-of-the-art methods. The code and hyper-parameter settings are released for reproducibility\footnote{\url{https://github.com/xiaolin1207/HTTN-master}}.
\end{abstract}
\section{Introduction}
\begin{figure}[t]
\subfigure[Label distribution in AAPD ]{
\begin{minipage}[t]{0.48\linewidth}
\centering
\includegraphics[width=4.13cm]{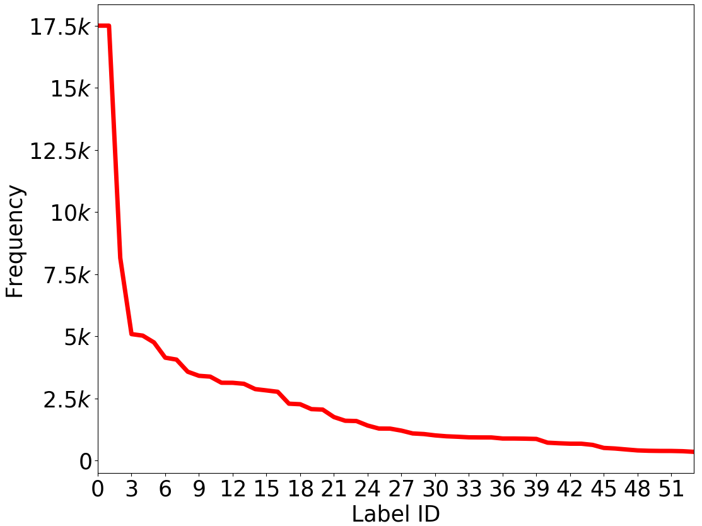}
%\caption{fig1}
\end{minipage}%
}%
\subfigure[Label distribution in RCV1 ]{
\begin{minipage}[t]{0.48\linewidth}
\centering
\includegraphics[width=4.1cm]{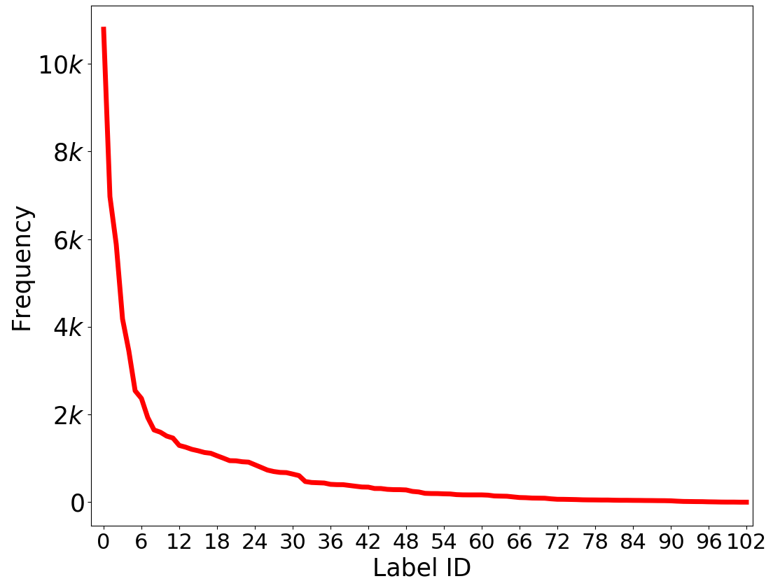}
%\caption{fig2}
\end{minipage}%
}%
\\
\subfigure[Label correlation in AAPD ]{
\begin{minipage}[t]{0.48\linewidth}
\centering
\includegraphics[width=4.1cm]{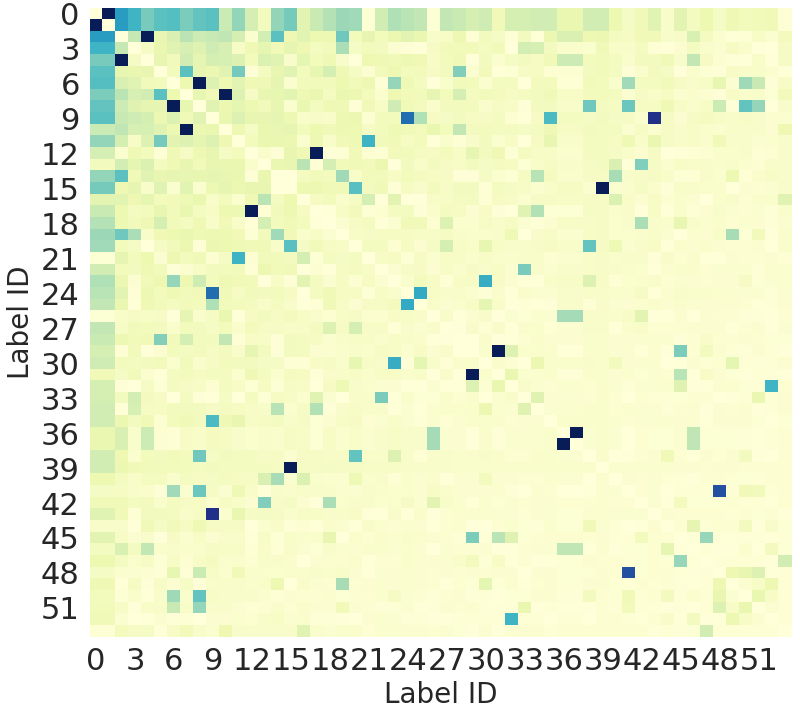}
%\caption{fig1}
\end{minipage}%
}%
\subfigure[Label correlation in RCV1]{
\begin{minipage}[t]{0.48\linewidth}
\centering
\includegraphics[width=4.2cm]{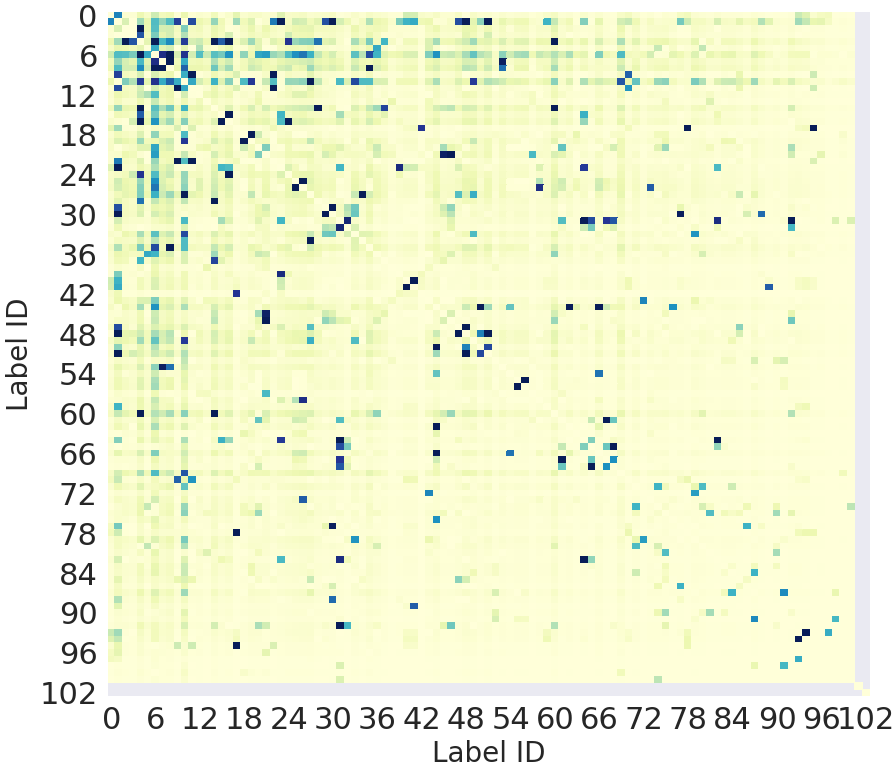}
%\caption{fig2}
\end{minipage}%
}%
\caption{The label distribution and pairwise label correlation coefficients in AAPD and RCV1 datasets. Here, labels are ordered by their frequency from the highest to the lowest in all subfigures. The points with bright color indicate higher correlation between labels in subfigure (c)-(d).}
\label{Fig:correlation}
\end{figure}

Multi-label text classification has become one of the core tasks in natural language processing and has been widely applied in topic recognition~\cite{yang2016hierarchical}, question answering~\cite{kumar2016ask}, sentimental analysis~\cite{cambria2014senticnet} and so on.
Even though various techniques have been proposed for multi-label learning, it is still a challenging task due to two main characteristics.
One important statistical characteristic is that multi-label data usually follows a paw-law distribution (called as long-tailed), especially for data with large number of labels.
As shown in Figure~\ref{Fig:correlation} (a)-(b),a few labels are associated with a large number of  documents (a.k.a. \emph{head} labels), while a large fraction of labels are associated with a small number of documents (a.k.a. \emph{tail} labels).
In this situation, learning classifiers for tail labels is much more difficult than that for head labels, due to the poor generalizability caused by insufficient training instances.

The other is label dependency because multiple labels may be assigned to one document, which makes it hard to separate classes with common instances.
On the other hand, fortunately, such label dependency is helpful to construct the correlation among labels for knowledge transferring.
Figure~\ref{Fig:correlation} (c)-(d) illustrate the pairwise label correlation on AAPD and RCV1 datasets. An interesting point is that there are strong correlations between head labels, as well as between head and tail labels.
To handle such multi-label data, an intuitive idea is to make use of these label dependency, which has been a hot research topic in multi-label learning. Most existing studies focus on how to exploit label structure \cite{zhang2018deep,huang2019label}, label content meaning \cite{pappas2019gile,xiao2019label,du2019explicit}, or label co-occurrence patterns~\cite{liu2017deep, kurata2016improved} to built one classifier on all labels. Recently, \cite{wei2019does} empirically demonstrate that tail labels impact much less than head labels on the whole prediction performance.
We are thus inspired to investigate an interesting and important question: \emph{Do the head labels help model construction on tail labels and further leverage long-tailed multi-label classification}?

Due to the long-tailed distribution, as we known, training head labels and tailed labels together will make the head labels dominate the learning procedure, which will sacrifice the prediction precision of head labels and recall of tail labels. To solve this issue, great effort has been done by designing proper instance sampling strategies (e.g., oversampling on tail label or undersampling on head label), constructing complex objective functions (e.g., label-aware margin loss, class-balanced loss) \cite{cao2019learning,cui2019class}, or transferring knowledge from head labels to tail labels \cite{liu2019large}.
These methods are proposed for long-tailed multi-class problems, however, long-tailed multi-label problems have barely been studied.
Recently, \citeauthor{macavaney2020interaction} \shortcite{macavaney2020interaction} exploit the extra information (label semantic embedding) to leverage the long-tailed multi-label classification. \citeauthor{yuan2019long} \shortcite{yuan2019long} train classifiers for head labels and tail labels separately.
Although they obtain impressive performance, the former suffers from high computing complexity and is limited by the existence of extra information; the later obviously ignores the correlation between head and tail labels, which has been proven important for multi-label learning.

In this paper, thus, we propose a \emph{Head-to-Tail Network} (HTTN) for long-tailed multi-label classification task. Its main idea is to take advantage of sufficient information among head labels and label dependency between head labels and tail labels. HTTN consists of two main parts. The first part aims to learn the meta-knowledge between the learning model on few-shot data and that on many-shot data, which is implemented with the aid of data-rich head labels. The second part tries to leverage the classifier construction on tail labels by transferring meta-knowledge learnt from head labels and exploiting label dependency between head and tail labels.
A good characteristic of HTTN is that only the classifier on head labels has to be trained, while the learning model on tail labels can be directly computed with the aid of meta-knowledge and their own instances. This strategy has a great by-product: once a new label with a few instances is coming, we need not retrain the whole model.
Meanwhile, to improve the robustness and generalization of tail label learning, an ensemble mechanism is designed for tail label prediction.
We summarize our main contributions as follows:
\begin{itemize}
	\item A head-to-tail network (HTTN) is proposed to tackle the long tail problem in multi-label text classification.
	\item HTTN effectively detects model transformation strategy (denoted as meta-knowledge about learning) from few-shot learning to many-shot learning on head labels.
    \item HTTN efficiently builds classifier for tail label with the aid of meta-knowledge and label dependency between head and tail labels.
    \item HTTN obtains promising performance on three widely-used benchmark datasets by comparing with several popular baselines.
\end{itemize}
The rest of the paper is organized via four Sections. Section 2 discusses the related work.  Section 3 describes the proposed HTTN model for multi-label text classification. The experimental setting and results are discussed in Section 4. A brief conclusion and future work are given in Section 5.
\section{Related work}
\textbf{Multi-label text classification (MLTC).}
In the line of MLTC, it has been a common knowledge  to  explore the label correlation.
\citeauthor{kurata2016improved} \shortcite{kurata2016improved} adopted label co-occurrence to initialize the final hidden layer  of the classifier network.
\citeauthor{zhang2018deep} \shortcite{zhang2018deep} learn the label embedding from the label co-occurrence graph to supervise the classifier construction.
In~\cite{du2019explicit}, the text-based label embedding is introduced to mine the fine-grained word-level classification clues.
A joint input-label embedding model was proposed in~\cite{pappas2019gile} to capture the structure of the labels and input documents and the interactions between the two.
Although multi-label learning benefits from the exploration of label correlation, these methods cannot well handle long tail problem because they treat all labels equally. In this case, the whole learning model will be dominated by the head labels.
Due to the insufficiency of training instances, in fact, tail labels were disclosed having much less impact than head labels on the whole prediction performance~\cite{wei2019does}.
The  language descriptions of labels were used in
\cite{macavaney2020interaction} to design a soft n-gram interaction matching model for non-frequent labels.
To treat head and tail labels in different manners, \cite{yuan2019long} proposed a two-stage and ensemble learning approach.
Although these methods obtain impressive results, they have limitations, such as high computing complexity, dependency on extra label information, ignorance of the label dependency among head and tail labels. \\
\textbf{Imbalanced data classification.} Another stream of work attacking the long tail problem is imbalanced learning, because the numbers of instances in head labels and tail labels have a big variance.
The main strategies  are discussed here. \\
\textit{Class distribution re-balancing strategy}:
The most popular ideas include   under-sampling the head classes~\cite{byrd2019effect},   over-sampling the tail classes~\cite{chawla2002smote,buda2018systematic,byrd2019effect}, and allocating large weights to tail classes in loss functions~\cite{cui2019class}.
Unfortunately, ~\cite{zhou2020bbn} and ~\cite{kang2019decoupling} empirically prove that re-balancing methods may hurt feature learning to some extent.
Recently, ~\cite{zhou2020bbn} proposed a unified Bilateral-Branch Network (BBN) integrating “conventional learning branch” and “re-balancing branch”. The former branch is equipped with the typical uniform sampler to learn the universal patterns for recognition, while the later branch is coupled with a reversed sampler to model the tail data.
A similar idea is used in ~\cite{kang2019decoupling} to decouple the learning procedure into representation learning and classification.
\\
\textit{Low-shot learning strategy}:
The low-shot learning shares similar features to the long-tail learning, because they both contain some labels with many instances, while the other labels have only few instances.
Low-shot learning aims to construct classifiers for data-poor classes with the aid of data-rich classes~\cite{hariharan2017low,gidaris2018dynamic,qi2018low}.
Among them, ~\cite{hariharan2017low} generate synthetic instances based on the head classifier, and incorporate them to train tail label learning model.
\citet{gidaris2018dynamic} proposed an attention based few-shot classification weight generator with the aid of base categories (i.e., head labels) to design the classifier, which can generalize better on ``unseen” categories and retain the patterns trained from base categories.
Similarly, based on a learner initially trained on base classes with abundant samples, a simple imprinting strategy was proposed in ~\cite{qi2018low} to effortlessly learn the novel categories with few samples.
\\
\textit{Knowledge transfer strategy}:
Another way to handle imbalanced data is to transfer the knowledge learned from data-rich classes  to help data-poor classes.
For example, \citet{wang2017learning} propose a meta-network on the space of model parameters learnt from head classes to determine the meta knowledge, which can be transferred to tail classes in a progressive manner.
In~\cite{liu2019large}, a dynamic meta-embedding method is proposed to learn both direct feature for few-shot categories and memory feature with the aid of many-shot categories, which can handle tail recognition robustness and open recognition sensitivity.

Our proposed method HTTN also adopts the knowledge transfer strategy, but focuses on the multi-label learning problem,  differing  from the existing methods working from the multi-class problem. In addition, HTTN builds generalized tail classifiers  directly from the  meta-knowledge  and  their  own  instances, and thus has more flexibility on processing the tail and novel rare classes.

\section{Proposed HTTN Method}
\textbf{Problem Definition:} Let $D = \{(x_{i},y_{i})\}_{i=1}^{N}$ denote the set of documents, which consists of $N$ documents with corresponding labels $Y=\{y_{i}\in \{0,1\}^{l}\}$, here $l$ is the total number of labels. Each document contains a sequence of words, $x_i=\{w_{1},\cdots,w_{q},\cdots,w_{n}\}$, where
$w_{q}\in \mathbb{R}^{k}$ is the $q$-th word vector (e.g., encoded by word2vec~\cite{pennington2014glove}).
Multi-label text classification (MLTC) aims to learn a classifier from $D$, which can
assign the most relevant labels to the new given documents. In this study, we divide  the label set into two parts: \emph{head} labels that are associated with many documents, and \emph{tail} labels that are associated with few documents.

The number of head labels and tail labels are $l_{head}$ and $l_{tail}$, respectively. The corresponding documents associated with head labels and tail labels form $D_{head}$ and $D_{tail}$, respectively. There may be overlapping documents between them. If a document belongs to both head   and tail label, it  appears in both $D_{head}$ and $D_{tail}$.
\begin{figure}[htbp]
    \centering
    \includegraphics[width=8cm]{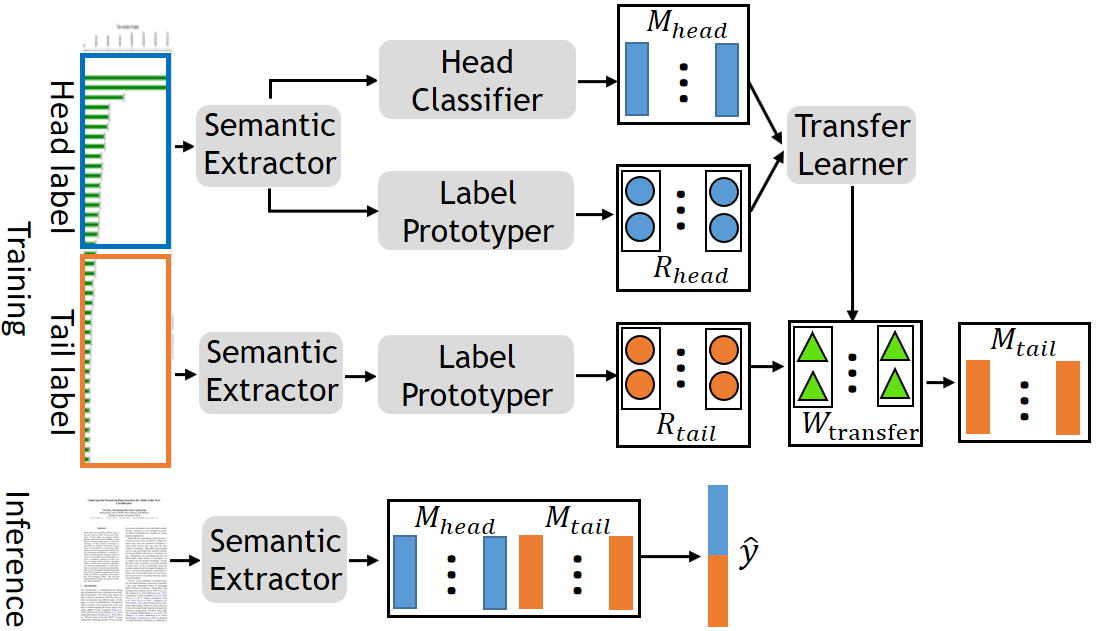}
    \setlength{\abovecaptionskip}{0.cm}
    \setlength{\belowcaptionskip}{-0.cm}
    \caption{The architecture of HTTN}
	\label{Fig:model}
\end{figure}
The framework of the proposed Head-to-Tail Network (HTTN) is shown in Figure~\ref{Fig:model}.
HTTN consists of three stages in the training process. Firstly, learning the head label classifiers by using a semantic extractor $\phi$, which extracts the semantic information from the documents by a Bi-LSTM with attention mechanism. The classifier weights of head labels are then learned, denoted as $M_{head}$. Secondly, a label prototyper generates a  prototype for each head label by  sampling the documents representation learned through $\phi$ and then taking average of these samples with the same label, denoted by $R_{head}$.
Thirdly, a transfer learner distils class-irrelevant  meta-knowledge $W_{transfer}$ for mapping
$R_{head}$ to $M_{head}$.
By the learned generic meta-knowledge $W_{transfer}$, the classifier weights of tail labels $M_{tail}$ can then be  inferred from their corresponding label prototype $R_{tail}$, although only few documents are available in tail labels.
The inference is done by sending given documents to the shared semantic extractor $\phi$, and then predicting the labels by the integration of $M_{head}$ and $M_{tail}$.
We next discuss the training process in details.

\subsection{Learning Head Label Classifiers}
\noindent \textbf{Semantic Extractor:}
We adopt the bidirectional long short-term memory (Bi-LSTM)~\cite{zhou2016text} with self-attention mechanism~\cite{tan2018deep} to learn the document representation. With the input word  $w_q$ in a document $x$, the hidden states of Bi-LSTM are updated as,
\begin{equation}\label{1}
\begin{split}
\overrightarrow{h_{q}} = LSTM(\overrightarrow{h_{q-1}},w_{q})\\
\overleftarrow{h_{q}} = LSTM(\overleftarrow{h_{q-1}},w_{q})
\end{split}
\end{equation}
where the hidden states $\overrightarrow{h_{q}},\overleftarrow{h_{q}}\in \mathbb{R}^{k}$ encode the forward and backward word context representations, respectively.
After taking all $n$ words, the whole document is represented as,
\begin{equation}\label{hlstm}
\centering
\begin{split}
H &=(\overrightarrow{H} ,\overleftarrow{H}) \; \in \mathbb{R}^{2k\times n} \\
\overrightarrow{H} &= (\overrightarrow{h_{1}},\overrightarrow{h_{2}},\cdots,\overrightarrow{h_{n}})\\
\overleftarrow{H} &= (\overleftarrow{h_{1}},\overleftarrow{h_{2}},\cdots,\overleftarrow{h_{n}})\\
\end{split}
\end{equation}
To intensify the representation with important words in each document, we adopt the attention mechanism~\cite{vaswani2017attention}, which has been successful used in various text mining tasks~\cite{tan2018deep,al2018hierarchical,you2018attentionxml},
\begin{equation}
\begin{split}
E &= softmax(W_{1}H)\\
r &= f(EH)=(EH^{T})W_{2}
\end{split}
\end{equation}
where $W_{1}\in \mathbb{R}^{1\times 2k}$ are the attention parameters, and $E\in \mathbb{R}^{1\times n}$  presents the contribution of all words to the document. The final document representation $r\in \mathbb{R}^{d}$ is obtained by a linear embedding layer $f(.)$ with parameter $W_{2}\in \mathbb{R}^{2k\times d}$. It is worth noting that the semantic extractor $r=\phi(x)$ is a shared block in the whole HTTN model.

\noindent \textbf{Head Classifier Construction:} Once having the document representation $r\in \mathbb{R}^{d}$, we can build the multi-label text classifier for head labels, e.g.,   by a one-layer neural network,
\begin{equation}
\begin{split}
\hat{y} = sigmoid(rM_{head})
\end{split}
\end{equation}where $M_{head}\in  \mathbb{R} ^{d\times l_{head}} $ is the weights to learn as head classifier parameters, and $l_{head}$ is the number of head labels. The sigmoid function  transfers the output values into   probabilities for assigning multiple labels to one document. Cross-entropy is thus used as  loss function,  whose suitability has been proved  for multi-label   learning~\cite{nam2014large},
\begin{equation}\label{5}
\begin{split}
\mathcal{L}_c=&-\sum^{l_{head}}_{j=1}\sum_{x_i \in D_{head}}(y_{ij}\log(\hat{y}_{ij}))\\
&+(1-y_{ij})\log(1-\hat{y}_{ij})
\end{split}
\end{equation}where $N$ is the number of training documents belonging to head, $l_{head}$ is the number of head label, $\hat{y}_{ij}\in [0,1]$ is the predicted probability, and $y_{ij} \in\{ 0,1\} $ indicates the ground truth of the $i$-th document along the $j$-th label. The head classifier weight $M_{head}$ can be learned by minimizing the above loss function.
\begin{table*}[t]
\small
\begin{center}\label{Tabel:dataset}
\caption{\label{font-table} Summary of Experimental Datasets.}
\begin{tabular}{c|cccccccc}
\hline  Datasets &  $N$ &  $M$ &  $D$ &  $L$ &  $\bar{L}$ &  $\tilde{L}$ &  $\bar{W}$ &  $\tilde{W}$\\ \hline
RCV1 & 23,149  & 781,265 & 47,236 & 103 & 3.18& 729.67 & 259.47 & 269.23\\
AAPD & 54,840  & 1,000 & 69,399 & 54 & 2.41 & 2444.04  & 163.42 & 171.65\\
EUR-Lex & 13,905  & 3,865 &33246 & 3,714 & 5.32& 19.93  & 1,217.47 & 1,242.13\\
\hline
\end{tabular}
\end{center}
\setlength{\abovecaptionskip}{0.cm}
	\setlength{\belowcaptionskip}{-0.cm}
{\small $N$ is the number of training instances, $M$ is the number of test instances, $D $ is the total number of words, $ L $ is the total number of classes, $\bar{L} $ is the average number of labels per document, $\tilde{L}$ is the average number of documents per label, $ \bar{W}$ is the average number of words per document in the training set, $ \tilde{W} $ is the average number of words per document in the testing set. }
\end{table*}
\subsection{Label Prototyper}
The label prototyper is designed to build a prototype for each class. We borrow the idea from meta-learning prototypical network \cite{snell2017prototypical}, which is an effective  multi-class few-shot classification approach.
For a head label $j$ (same later for a tail class), we sample $t$ documents and get their representation $\{r^{j}_{1},\cdots, r^{j}_{t}\}$. Then the prototype is obtained by taking average of these vectors,
\begin{equation}\label{one1}
p^{j}_{head}= avg\{r^{j}_{1},\cdots, r^{j}_{t}\}
\end{equation}In multi-class prototypical network \cite{snell2017prototypical}, one prototype is built for each class, and all prototypes are independent. However, our prototypes built here in multi-label learning are correlated, because the sampled documents  of one label can also be sampled for other
labels. The correlation between prototypes is consistent with the correlation between labels.
\subsection{Transfer Learner}
The transfer learner is designed to link the  (few-shot) label prototype $p^{j}$ and the corresponding many-shot classifier parameter  $m^j$. For head labels, we have obtained their many-shot classifier parameter $m^j_{head}\in M_{head}$, as well as their label prototype $p^j_{head}$. Therefore, a transfer function can be learned to map $p^j_{head}$ to $m^j_{head}, j=1...l_{head}$, by minimizing
\begin{equation}
%\begin{split}
\mathcal{L}_t=\sum_{j=1}^{l_{head}}||m_{head}^j-W_{transfer}p^{j}_{head}||^2
%\end{split}
\end{equation}where $W_{transfer}\in  \mathbb{R} ^{d\times d}$ is the parameter of the transfer learner. It captures the generic and class irrelevant transformation from few-shot label prototypes to many-short classifier parameters.
For each head label, we sample $S$  times to have different $p^j_{head}$ for training a generalizable transfer learner.
The $S$ usually is a small constant, e.g., 30 or 40. The sensitivity analysis of $S$ is presented in section 4.5.

Since the generic transfer learner maps the (few-shot) prototype to (many-shot) classifier parameters as a class-irrelevant transformation, we can use it to map the (few-shot) tail prototypes to their (many-shot) classifier parameters.
For a tail label $z$, we also sample $t$ documents and get their representation $\{r^{z}_{1},\cdots, r^{z}_{t}\}$ by the trained semantic extractor. Then, we use the label prototyper to get the prototype of the tail label,
\begin{equation}
p^{z}_{tail} = avg\{r^{z}_{1},\cdots, r^{z}_{t}\}.
\end{equation}Thereafter, the tail label classifier parameters are estimated  by using the transfer learner,
\begin{equation}
\hat{m}^z_{tail}=W_{transfer}p^{z}_{tail}.
\end{equation}
This $\hat{m}^z_{tail}$ is an estimation of a tail classifier when it has  many-shot document instances. As discussed before, one of the most important characteristics of MLTC is the label correlation caused by label co-occurrence. Although the label prototyper can capture the label co-occurrence by making a same document instance contributing to more than one label prototype since the document may have multiple labels, the label correlation has not been sufficient explored due to the random sampling process. Especially in the tail labels, fully considering the correlation between tail labels and head labels can effectively improve the classification performance on the tail labels.
We thus propose a tail label attention module, which aims to enhance the tail label classifiers by exploring their correlation with the head labels.
For each tail label $p^{z}_{tail}$, we calculate the attention score between it and each head prototype $p^{j}_{head}$:
\begin{equation}
\begin{split}
e_{zj}  &=f_{att}(p^{z}_{tail},p^{j}_{head})\\
 \alpha_{zj} & = softmax(e_{zj})=\frac{exp(e_{zj})}{\sum_{k=1}^{l_{head}}exp(e_{jk})}\\
p^{z}_{att}  &  = \sum_{j}(\alpha_{zj}p^{j}_{head})\\
p^{z}_{new}& =  avg(p^{z}_{att},p^{z}_{tail}).
\end{split}
\end{equation}
Then, the same transfer learner is applied to estimate the tail label classifier parameters,
\begin{equation}
\hat{m}_{tail}^z=W_{transfer}p^{z}_{new}\\
\end{equation}
which is concatenated with the head label classifier, forming  the whole classifier  for inference:
\begin{equation}
M= cat[M_{head}:\hat{M}_{tail}].
\end{equation}Given a testing document, it will first go through the semantic extractor $\phi$ to have its representation vector $r$, and then get the predicted label by $\hat{y}=sigmoid(rM)$.\\
\noindent \textbf{Ensemble HTTN:}
In early research, ensembles were proven empirically and theoretically to possess better performance than any single component. Hence, to improve the robustness of the classification process, we extend HTTN in an ensemble way. Ensemble HTTN (EHTTN) is designed to increase the accuracy of a single classifier by building several different tail classifiers. In summary, we sample the documents belonging to tail labels $G$ times, and use the transfer learner to obtain multiple classifier weights $\{\hat{M}^{1}_{tail} ,\cdots, \hat{M}^{G}_{tail}\}$ for tail labels, and thus multiple $\{M_{1},\cdots, M_{G}\}$ used for inference.
Ensemble HTTN has the following advantages: 1) Robustness. If only sampling once, the model will be greatly affected by the quality of the randomly sampled documents. However, ensemble HTTN can avoid the caused problems. 2) Flexibility.  Ensemble HTTN is flexible on handling tail labels with different number of instances. Even in the long-tail part, some tail labels have dozens of instances, while the others have a few instances. Using a single batch of a fixed number of instances may under-sample the former and leave  the latter out.% 3)Diversity
\begin{table*}[t]
\footnotesize
\setlength{\abovecaptionskip}{0.cm}
\setlength{\belowcaptionskip}{-0.cm}
\begin{center}
\caption{\label{font-table1} Comparing HTTN with baselines on AAPD dataset.}
\begin{tabular}{|cc|ccc|cc|c|}
\hline
 \multicolumn{2}{|c|}{Method}&P@1&P@3&P@5&nDCG@3&nDCG@5&F1-score\\
\hline
\multicolumn{2}{|c|}{Joint}&78.20&55.21&37.89&73.42&77.63&63.88\\
\hline
\multicolumn{2}{|c|}{DXML}&80.54&56.30&39.16&77.23&80.99&65.13\\
\hline
\multicolumn{2}{|c|}{XML-CNN}&74.38&53.84&37.79&71.12&75.93&65.35\\
\hline
\multicolumn{2}{|c|}{OLTR}&78.96&56.28&38.60&74.66&78.58&62.48\\
\hline
\multicolumn{2}{|c|}{Imprinting}&68.68&38.22&23.71&55.30&55.67&25.58\\
\hline
\multicolumn{2}{|c|}{BBN}&81.56&57.81&39.10&76.92&80.06&66.73\\
\hline
$l_{tail}$=18&LTMCP&78.12&55.19&37.67&75.18&75.43&62.84\\
\hline
$l_{tail}$=18&HTTN&82.04&57.12&39.33&76.98&80.69&67.71\\
$l_{tail}$=18 &EHTTN&83.34&59.06&40.30&77.75&81.65&68.84\\
\hline
$l_{tail}$=9&LTMCP&78.51&56.02&38.46&75.19&76.05&63.59\\
\hline
$l_{tail}$=9&HTTN&82.49&58.72&40.31&78.20&81.24&68.14\\
$l_{tail}$=9&EHTTN& \bf83.84&\bf 59.92& \bf40.79& \bf79.27&  \bf82.67& \bf69.25\\
\hline
\end{tabular}
\end{center}
\setlength{\abovecaptionskip}{0.cm}
\setlength{\belowcaptionskip}{-0.cm}
\begin{center}
\caption{\label{font-table2} Comparing HTTN with baselines on RCV1 dataset.}
\begin{tabular}{|cc|ccc|cc|c|}
\hline
\multicolumn{2}{|c|}{Method}&P@1&P@3&P@5&nDCG@3&nDCG@5&F1-score  \\
\hline
\multicolumn{2}{|c|}{Joint}&92.18&72.33&47.35&83.02&81.47&75.19\\
\hline
\multicolumn{2}{|c|}{DXML}&94.04&78.65&54.38&89.83&90.21&75.76\\
\hline
\multicolumn{2}{|c|}{XML-CNN}&95.75&78.63&54.94&\bf89.89&90.77&75.92\\
\hline
\multicolumn{2}{|c|}{OLTR}&93.79&61.36&44.78&74.37&77.05&56.44\\
\hline
\multicolumn{2}{|c|}{Imprinting}&77.38&47.96&31.45&58.83&57.91&26.35\\
\hline
\multicolumn{2}{|c|}{BBN}&94.61&77.98&54.25&88.97&89.68&\bf78.65\\
\hline
$l_{tail}$=28&LTMCP&90.47&74.57&51.59&85.31&85.83&73.99\\
\hline
$l_{tail}$=28&HTTN&94.11&75.92&52.85&87.02&87.98&76.09\\
$l_{tail}$=28&EHTTN&95.62&77.25&54.28&87.46&88.46&76.92\\
\hline
$l_{tail}$=14&LTMCP&91.39&73.04&49.76&83.30&83.93&74.67\\
\hline
$l_{tail}$=14&HTTN&94.70&77.83&54.21&88.49&89.05&76.86\\
$l_{tail}$=14&EHTTN& \bf95.86& \bf78.92&\bf55.27& 89.61& \bf90.86& 77.72\\
\hline
\end{tabular}
\end{center}
\setlength{\abovecaptionskip}{0.cm}
\setlength{\belowcaptionskip}{-0.cm}
\begin{center}
\caption{\label{font-table3}Comparing HTTN with baselines on EUR-Lex dataset.}
\begin{tabular}{|cc|ccc|cc|c|}
\hline
\multicolumn{2}{|c|}{Method}&P@1&P@3&P@5&nDCG@3&nDCG@5&F1-score \\
\hline
\multicolumn{2}{|c|}{Joint}&79.04&64.89&55.00&69.20&63.60&52.51\\
\hline
\multicolumn{2}{|c|}{DXML}&80.41&66.74&56.33&70.03&63.18&53.28\\
\hline
\multicolumn{2}{|c|}{XML-CNN}&78.20&65.93&53.81&68.41&60.54&51.98\\
\hline
\multicolumn{2}{|c|}{OLTR}&65.62&52.34& 42.69&55.73&50.57&22.64\\
\hline
\multicolumn{2}{|c|}{Imprinting}&62.16&40.25&29.07&45.46&38.24&9.94\\
\hline
\multicolumn{2}{|c|}{BBN}&76.22&60.40&49.45&64.26&58.54&41.01\\
\hline
$l_{tail}$=1238&LTMCP&75.23&60.12&49.36&64.89&58.23&48.10\\
$l_{tail}$=768&LTMCP&77.26&62.39&52.10&67.18&60.54&50.33\\
\hline
$l_{tail}$=1238&HTTN& 80.53& 66.96& 55.71 &70.35& 63.87& 53.44\\
$l_{tail}$=768&HTTN& \bf81.14& \bf67.62& \bf56.38 &\bf70.89& \bf64.42& \bf53.72\\
\hline
\end{tabular}
\end{center}
\setlength{\abovecaptionskip}{0.cm}
\setlength{\belowcaptionskip}{-0.cm}
\end{table*}
\section{Experiments}
In this section, we evaluate the proposed model on three datasets
by comparing with the state-of-the-art methods in terms of widely used metrics, P@K and nDCG@K (k=1,3,5) and F1-score.
\subsection{Experimental Setting}
\textbf{Datasets:}
Three multi-label text datasets are  used to evaluate the HTTN model, AAPD, RCV1 and EUR-Lex. Their label distributions all follow the power-low distribution, as shown in Figure~\ref{Fig:correlation} (the label distribution of EUR-Lex is presented in the supplementary document due to the space limit).
The  benchmark datasets have defined the training and testing split. We follow  the same  data usage for all evaluated models. The datasets are  summarized in Table \ref{font-table}.\\
\textbf{Baseline Models:} To demonstrate the effectiveness of HTTN on the benchmark datasets,
we selected the seven most representative baseline models in the different groups of related work discussed in the second session.
\begin{itemize}
\vspace{-0.1em}
\item {\bf Joint:} it uses Bi-LSTM with self-attention mechanism to tackle multi-label text classification without differentiating the head and tail labels, i.e., learning the classifier for them in a joint way.
	\vspace{-0.2em}
\item {\bf XML-CNN}~\cite{liu2017deep}: it adopts Convolutional Neural Network (CNN) and a dynamic pooling technique to extract high-level feature for large-scale multi-label text classification.
	\vspace{-0.2em}
\item {\bf DXML}~\cite{zhang2018deep}: it tries to solve the multi-label long tail problem by considering the label structure from the label co-occurrence graph.
	\vspace{-0.2em}
\item {\bf LTMCP}~\cite{yuan2019long}: it introduces an ensemble method to tackle long-tailed multi-label training. The DNN and linear classifier are combined to deal with the head label and tail label respectively.
	\vspace{-0.2em}
\item {\bf BBN}~\cite{zhou2020bbn}: it  takes care of both representation learning and classifier learning for exhaustively improving the performance of long-tailed tasks.
	\vspace{-0.2em}
\item {\bf Imprinting}~\cite{qi2018low}: it  computes embeddings of novel examples and set novel weights in the final layer directly.
	\vspace{-0.2em}
\item {\bf OLTR}~\cite{liu2019large}: it learns dynamic meta-embedding in order to share visual knowledge between head and tail classes.
\end{itemize}
\textbf{Parameter Setting:}
For all three datasets, we use Glove~\cite{pennington2014glove} to get the word embedding in 300-dim. LSTM hidden state dimension $k$ is set to 300. The parameter  $d=128$ for $W_2$ and $W_{transfer}$. The number of sampled instances $t$ for label prototyper in  AAPD, RCV1 and EUR-Lex are $t=5,5,1$, respectively. The whole model is trained via Adam~\cite{kingma2014adam} with the learning rate being 0.001.
  AAPD and RCV1    have   54 and 103 labels, respectively. To test the performance on different number of tail labels, we set $l_{tail}$=18 and 9  in AAPD, and   $l_{tail}$=28 and 14 in RCV1.
 For EUR-Lex dataset,  we select the last 768 one-shot tail labels and  the 1238 less than three-shot tail labels.
For the ensemble HTTN, we set   $G=30$ for  AAPD and RCV1.  $G=1$ for EUR-Lex, because there are many one-shot labels in EUR-Lex. %, and present the sensitivity analysis of $G$ later.
We used the default parameters for the DXML, XML-CNN, EXAM, and LTMCP models. The baselines OLTR, Imprinting, and BBN deal with the long tail problem on the image recognition, the feature extractor used was the ResNet-10,  ResNet-32 and others. For a fair comparison, we replace the feature extractor with Bi-LSTM with attention.
%All experiment results reported in this paper are averaged from five independent runs.
The parameters of all baselines are either adopted from their original papers or determined by experiments.
\subsection{Results Comparison and Discussion}
The results on three datasets are presented
in Table \ref{font-table1}, Table \ref{font-table2}, and Table \ref{font-table3}.
The best results are marked in bold.
From Table \ref{font-table1} to Table \ref{font-table3}, we can make a number of observations. Firstly, \emph{Imprinting} is worse than other  methods because it only copies the embedding activations for a novel exemplar as the new set of classifier parameters.
DXML explores the label correlation by  the label graph to alleviate the long tail problem in MLTC, so they can get the satisfying results. OLTR learns the dynamic meta-embedding to help the tail label classification. LTMXP combines linear model and DNN to train the documents belonging to tail label and head label respectively. However, OLTR and LTMXP both don't consider the correlation between the head labels and tail labels, which is in fact important for long tail MLTC task. The \emph{Joint} method  trained the documents belonging to the head labels and the tail labels jointly together, resulting in good results on the head labels, but bad results on the tail labels. The EHTTN transfers the meta-knowledge from the head labels to tail labels, and ensembles multiple sampled documents from tail labels to further improve the robustness of tail label classification. The results demonstrate the superiority of the proposed EHTTN on all metrics  for MLTC. In EUR-Lex training set, there are 768 labels with only one training document. So $l_{tail}=768$ is a setting equivalent to one-shot learning. The high data scarcity causes several methods have poor performance. Especially in OLTR method that borrows information from the learned memory to help tail label classification,  it cannot get a comprehensive memory to help tail label classification, because  each tail label has only one document. HTTN in this one-shot setting outperform other methods on all measures. It is also interesting to find in all three datasets that HTTN/EHTTN performs better when $l_{tail}$ is smaller. The reason is that when   $l_{tail}$ is smaller, more head labels are used for distilling $W_{transfer}$ to have richer meta-knowledge. We present the detailed analysis of the impact of $l_{tail}$ in the supplementary document.
The results in Table  \ref{font-table1},   \ref{font-table2}, and  \ref{font-table3}
answer the question we had: \emph{the head labels do  help a lot for long-tailed multi-label text classification}. In addition, \emph{more head labels can be more helpful on learning the meta-knowledge}.
\begin{figure}[t]
\subfigure[AAPD]{
\begin{minipage}[t]{0.48\linewidth}
\centering
\includegraphics[width=4.1cm]{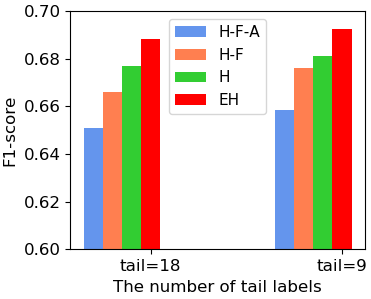}
%\caption{fig1}
\end{minipage}%
}%
\subfigure [RCV1 ]{
\begin{minipage}[t]{0.48\linewidth}
\centering
\includegraphics[width=4.1cm]{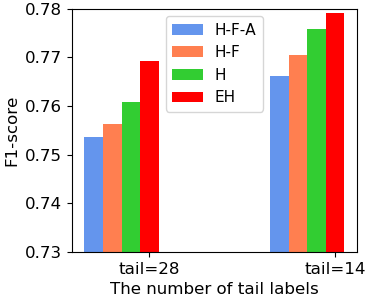}
%\caption{fig2}
\end{minipage}%
}%
\caption{ Ablation test on two datasets.}
\label{Fig:ablation}
\setlength{\abovecaptionskip}{0.cm}
\setlength{\belowcaptionskip}{-0.cm}
\end{figure}
\begin{figure}[t]
	\centering
	\includegraphics[width=7.5cm]{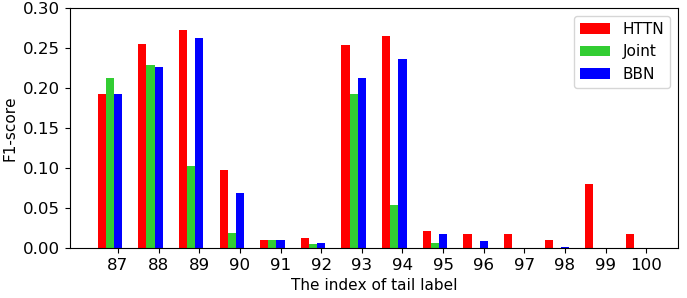}
	\caption{F1-score on RCV1 tail labels.}
	\label{Fig:tail}
\setlength{\abovecaptionskip}{0.cm}
\setlength{\belowcaptionskip}{-0.cm}
\end{figure}

\begin{figure}[t]
	\centering
	\includegraphics[width=4cm]{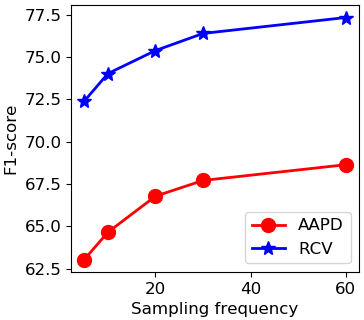}
	\caption{Sensitivity to the sampling frequency.}
	\label{Fig:sensi}
\setlength{\abovecaptionskip}{0.cm}
\setlength{\belowcaptionskip}{-0.cm}
\end{figure}
\subsection{Ablation Test}
An ablation test would provide informative analysis about the effect of  different components of the
proposed HTTN, which can be taken apart as HTTN without attention modular and fine-tuning (denoted as H$-$F$-$A), HTTN without fine-tuning  (H$-$F) but with attention, the complete HTTN (H), and the ensemble of multiple HTTN (denoted as EH) in Figure \ref{Fig:ablation}.  The results were obtained on AAPD and RCV1 datasets.
There are two interesting observations: 1) It is always preferable to use the ensemble strategy, as shown by the superior performance of {  EH}; and 2) The result of {  H$-$F} is always better than {  H$-$F$-$A}, because the attention modular is designed to explore the correlation between the head and tail labels, thus
improves the classification performance. {H} is better than {  H$-$F$-$A} and  {H$-$F}, indicating the fine-tuning can further improve classification performance.
\subsection{Performance analysis on tail labels}
To further verify the proposed HTTN, we compare it with \emph{Joint} and BBN~\cite{zhou2020bbn} on only tail labels. Figure \ref{Fig:tail} shows their  F1-score  on the tail labels in RCV1.
We can see that the F1-scores of HTTN on most of the tail labels are higher than those from the \emph{Joint} and BBN model, especially on the extreme tail label 97, 98, 99 and 100. The number of documents belonging to them is 6, 3, 2, 2, respectively. Due to the high data scarcity, the results predicted by \emph{Joint} and BNN on them are all 0 (no positive).
The meta-knowledge learned in HTTN does help to build effective tail label classifiers, making non-zero positive prediction.
\subsection{Sensitivity of $S$ in transfer learner}
For investigating the impact of the number of sampling frequency $S$ in transfer learner, we vary $S$ from 5 to 60, and show its influence on F1-score in Figure \ref{Fig:sensi}. Increasing $S$ from 5 to 20 can greatly help HTTN to gain strong improvement in both datasets. That's to say, sampling multiple prototypes in head labels can effectively strengthen the generalizability of $W_{transfer}$ and distil the  class irrelevant meta-knowledge to transfer..
\subsection{Analysis of ensemble HTTN}
\begin{figure}[t]
\subfigure[AAPD]{
\begin{minipage}[t]{0.48\linewidth}
\centering
\includegraphics[width=4.1cm]{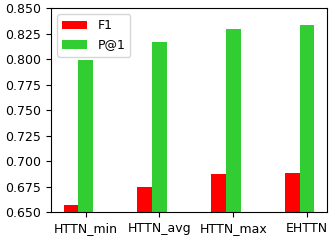}
\setlength{\abovecaptionskip}{0.cm}
\setlength{\belowcaptionskip}{-0.cm}
%\caption{fig1}
\end{minipage}%
}%
\subfigure [RCV1 ]{
\begin{minipage}[t]{0.48\linewidth}
\centering
\includegraphics[width=4.1cm]{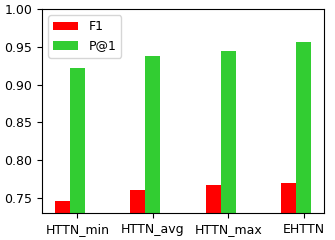}
\setlength{\abovecaptionskip}{0.cm}
\setlength{\belowcaptionskip}{-0.cm}
%\caption{fig2}
\end{minipage}%
}%
\caption{ The HTTN members  in EHTTN.}
\label{Fig:EHTTN}
\setlength{\abovecaptionskip}{0.cm}
\setlength{\belowcaptionskip}{-0.cm}
\end{figure}

In order to further verify the robustness and effectiveness of  Ensemble HTTN, in Figure \ref{Fig:EHTTN}, we compare the results of EHTTN with that of the worst member (HTTN\_min), the best member (HTTN\_max) and the average results of all members (HTTN\_avg).
We can see that EHTTN always achieves the best. Learning only one  $\hat{M}_{tail} $ can result  in unstable performance,  depending on the quality of the sampled instances. Learning more  $\hat{M}_{tail} $ can strengthen the classifier with higher robustness and  diversity.

In summary, extensive experiments are carried out on three MLTC benchmark datasets with various scales. The results demonstrate that the proposed HTTN can achieve superior performance compared with the seven baselines. In particular, effectiveness of HTTN is shown on tail labels.
\section{Conclusions}
A Head-to-Tail Network (HTTN) is proposed in this paper for long tail multi-label text classification. By using the documents belonging to the head labels,  a transfer learner learns the meta-knowledge, which maps the class weights learned by few-shot to the class weights learned by many-shot. This generic and class-irrelevant meta-knowledge effectively improves the tail label classification performance. Extensive experiments on benchmark datasets demonstrate the superiority of HTTN, comparing with the state-of-the-art methods. With HTTN, the head labels do help for long-tailed multi-label text classification.
\section{ Acknowledgments}
This work was supported in part by the National Natural Science Foundation of China under Grant 61822601, 61773050, 61632004 and 61828302; The Beijing Natural Science Foundation under Grant Z180006; The National Key Research and Development Program of China under Grant 2020AAA0106800 and 2017YFC1703506; The Fundamental Research Funds for the Central Universities (2019JBZ110); And King Abdullah University of Science$\&$Technology, under award number FCC/1/1976-19-01.

\bibliographystyle{aaai21}
\bibliography{aaai21}

\end{document}